\documentclass[conference]{IEEEtran}
\IEEEoverridecommandlockouts

\usepackage{cite}
\usepackage{amsmath,amssymb,amsfonts}
\usepackage{graphicx}
\usepackage{textcomp}
\usepackage[table]{xcolor}
\usepackage{booktabs}
\usepackage{array}
\usepackage{subfig}
\usepackage{hyperref}

\definecolor{groupbg}{gray}{0.92}

\begin{document}

\title{SMILESGNN: Interpretable Clinical Toxicity Prediction via SMILES-Graph Cross-Attention Fusion}

\author{
\IEEEauthorblockN{Quang Minh Nguyen}
\IEEEauthorblockA{\textit{National Economics University}\\
Hanoi, Vietnam\\
11247324@st.neu.edu.vn}
\and
\IEEEauthorblockN{Thuy Quynh Nguyen}
\IEEEauthorblockA{\textit{National Economics University}\\
Hanoi, Vietnam\\
11247346@st.neu.edu.vn}
\and
\IEEEauthorblockN{Duc Minh Le}
\IEEEauthorblockA{\textit{National Economics University}\\
Hanoi, Vietnam\\
11247320@st.neu.edu.vn}
\and[\hfill\newline\hfill]
\IEEEauthorblockN{Ho Nhat Minh Nguyen}
\IEEEauthorblockA{\textit{National Economics University}\\
Hanoi, Vietnam\\
11247321@st.neu.edu.vn}
\and
\IEEEauthorblockN{Thanh Long Dai Doan}
\IEEEauthorblockA{\textit{VNU University of Science}\\
Hanoi, Vietnam\\
doandaithanhlong\_sdh21@hus.edu.vn}
\and
\IEEEauthorblockN{Trong Nghia Nguyen\thanks{Corresponding author: nghiant@neu.edu.vn}}
\IEEEauthorblockA{\textit{National Economics University}\\
Hanoi, Vietnam\\
nghiant@neu.edu.vn}
}

\maketitle

\begin{abstract}
Drug toxicity prediction is critical for reducing late-stage attrition in drug discovery, yet remains challenging due to severe class imbalance, scaffold-based generalization, and the clinical need for interpretable predictions. Single-modality approaches-SMILES Transformers or graph neural networks-capture complementary aspects of molecular structure, while sequence-only models cannot directly provide graph-attributed explanations. We present SMILESGNN, a multimodal architecture that fuses a SMILES Transformer encoder and a GATv2 graph encoder via cross-attention, and SMILESGNN-PT, a variant using a ChemBERTa-2 pretrained backbone. The design retains an explicit graph branch within the predictive pipeline, supporting GNNExplainer-based analysis of substructures associated with toxic predictions. On ClinTox, SMILESGNN achieves AUC-ROC 0.987\,\textpm\,0.012 and F1 0.906\,\textpm\,0.039 with only 0.4M parameters, performing competitively with a strong SMILES Transformer and a larger ChemBERTa-2/GATv2 concat-fusion baseline. On Tox21 (12 tasks), SMILESGNN-PT obtains mean AUC-ROC 0.750\,\textpm\,0.002, comparable to ChemBERTa-2 alone and the same-backbone concat-fusion baseline. Overall, the results suggest that cross-attention is a practical fusion alternative that preserves competitive predictive performance while enabling graph-based interpretability support.
\end{abstract}


\section{Introduction}
Drug discovery is a lengthy and costly process, with clinical development alone spanning over a decade~\cite{ref_seal2025}. A primary driver of late-stage attrition is unexpected toxicity: compounds that pass early screening may still fail in clinical trials due to organ toxicity or adverse drug reactions not apparent until human exposure. Early toxicity prediction from molecular structure is therefore critical, enabling researchers to prioritize safe candidates and reduce costly failures.

Three core challenges make clinical toxicity prediction particularly difficult. First, \textit{class imbalance}: the ClinTox benchmark~\cite{ref_moleculenet} has a 1:11.5 toxic-to-nontoxic ratio, causing models to collapse toward the majority class. Second, \textit{scaffold-based generalization}: models must predict toxicity for structurally novel compounds unseen during training, making scaffold-based evaluation essential. Third, \textit{multi-scale structural complexity}: toxicity arises from local features (reactive functional groups, halogenated rings) and global scaffold geometry simultaneously.

Deep learning has enabled powerful molecular representations, yet single-modality approaches — SMILES Transformers or GNNs — each capture only part of the structural picture. This motivates multimodal fusion. Beyond predictive accuracy, \textit{interpretability} is a distinct clinical requirement: toxicologists need to identify which atomic substructures drive a toxic classification. Sequence-only models cannot provide graph-attributed explanations; GNNExplainer~\cite{ref_gnnexplainer} requires an explicit graph pathway in the predictive model. Cross-attention fusion is intended to couple SMILES and graph representations so that post-hoc graph attributions can be analyzed for the GATv2 branch, while ablation results quantify how much each branch contributes. A further open question is whether cross-attention outperforms \textit{concatenation} — none of the existing methods address either requirement under the extreme class imbalance of clinical toxicity prediction.

To address these gaps, we propose SMILESGNN, a multimodal architecture fusing a SMILES Transformer encoder and a GATv2 graph encoder via cross-attention, trained with focal loss for class imbalance, and SMILESGNN-PT, its pretrained-backbone variant using ChemBERTa-2~\cite{ref_chemberta2}. Our contributions are: (1) the dual-pathway cross-attention architecture enables \textit{GNNExplainer-based graph attribution}, highlighting chemically interpretable toxicophore-like substructures-a capability sequence-only models cannot provide; (2) SMILESGNN achieves competitive AUC-ROC ($0.987\!\pm\!0.012$), outperforms all graph-based single-modality baselines, and exceeds the sequence-only baseline in minority-class F1 with only 0.4M parameters; (3) a controlled comparison under identical backbone conditions shows cross-attention is competitive with concatenation while retaining an explicit graph branch for interpretability; (4) SMILESGNN-PT generalizes to pretrained backbones, matching the same-backbone concat-fusion baseline on both ClinTox and Tox21 (12 tasks).

\section{Related Work}
Molecular fingerprints such as ECFP~\cite{ref_ecfp} remain widely used baselines but lose structural detail through fixed-length hashing. Graph Neural Networks (GNNs) operate directly on molecular graphs via message passing; prominent variants include GIN~\cite{ref_gin}, GATv2~\cite{ref_gatv2}, DMPNN~\cite{ref_dmpnn}, AttentiveFP~\cite{ref_attentivefp}, and the self-supervised pretrained GNN backbone of~\cite{ref_pretrained_gin}. On the sequence side, Transformer models~\cite{ref_molecular_transformer} treat SMILES strings as token sequences, with ChemBERTa-2~\cite{ref_chemberta2} and MoLFormer-XL~\cite{ref_molformer} demonstrating strong performance through large-scale SMILES pretraining~\cite{zhang2025artificial}.

Multimodal methods combine these complementary views. FP-GNN~\cite{ref_fpgnn} showed that concatenating ECFP fingerprints with GNN embeddings consistently outperforms either modality alone. More recently, Zang et al.~\cite{zang2025molecular} proposed gated fusion with representation decoupling, showing selective modality weighting outperforms fixed concatenation. However, none of these methods address extreme class imbalance, provide a controlled cross-attention versus concatenation comparison, or evaluate graph-aware post-hoc attribution under a multimodal toxicity setting.

\section{Methodology}
\label{sec:method}
\begin{figure*}[!t]
\centerline{\includegraphics[width=0.8\textwidth]{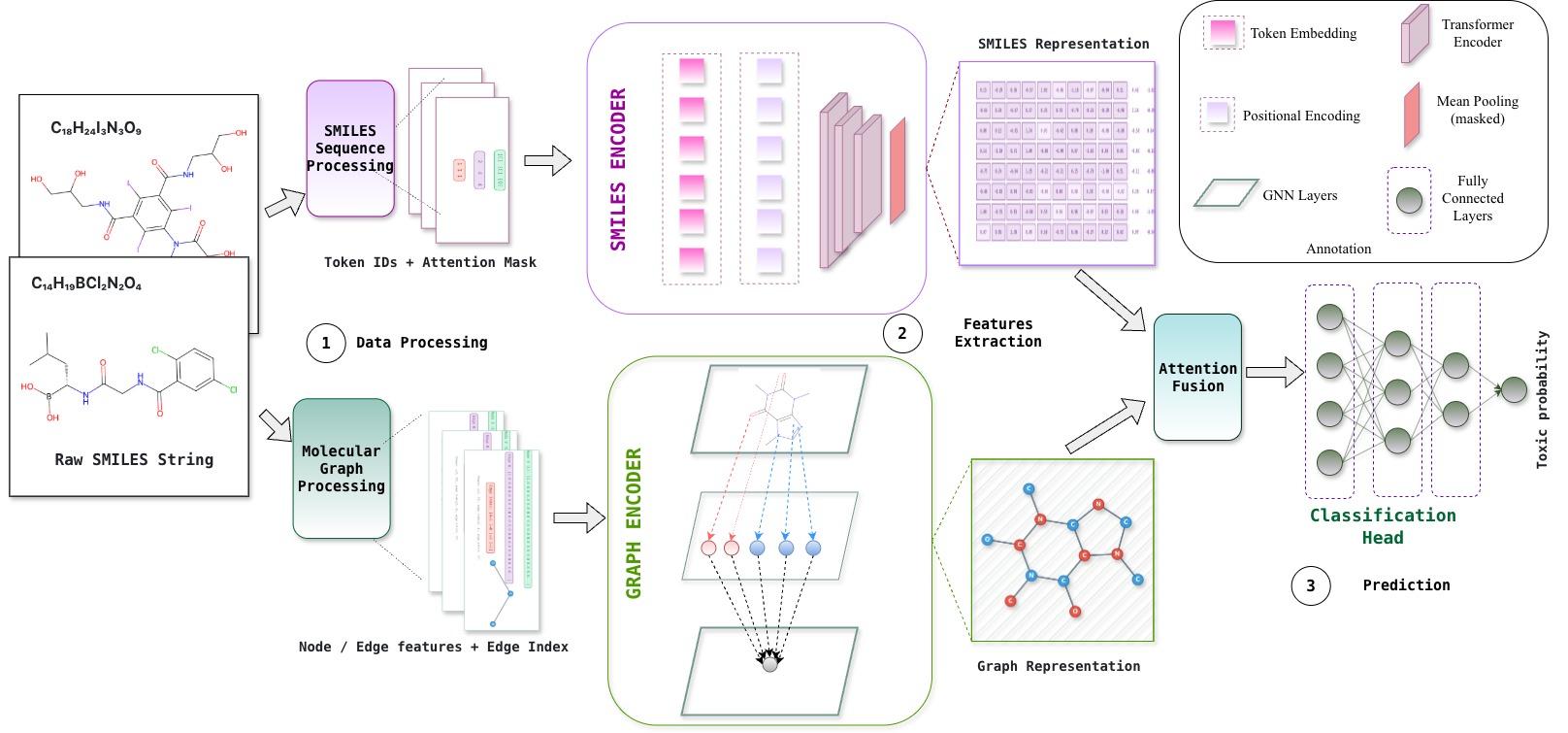}}
\caption{SMILESGNN architecture: (1) Transformer SMILES encoder; (2) GATv2 graph encoder; (3) cross-attention fusion module; (4) MLP predictor head. The dual-pathway design captures complementary sequence and graph structural information; cross-attention couples the GATv2 pathway with the SMILES pathway, supporting GNNExplainer-based graph interpretability.}
\label{fig:architecture}
\end{figure*}
\subsection{Dataset}
\begin{figure}[htbp]
\centering
\subfloat[Non-toxic (FDA-approved)]{\includegraphics[width=0.48\columnwidth]{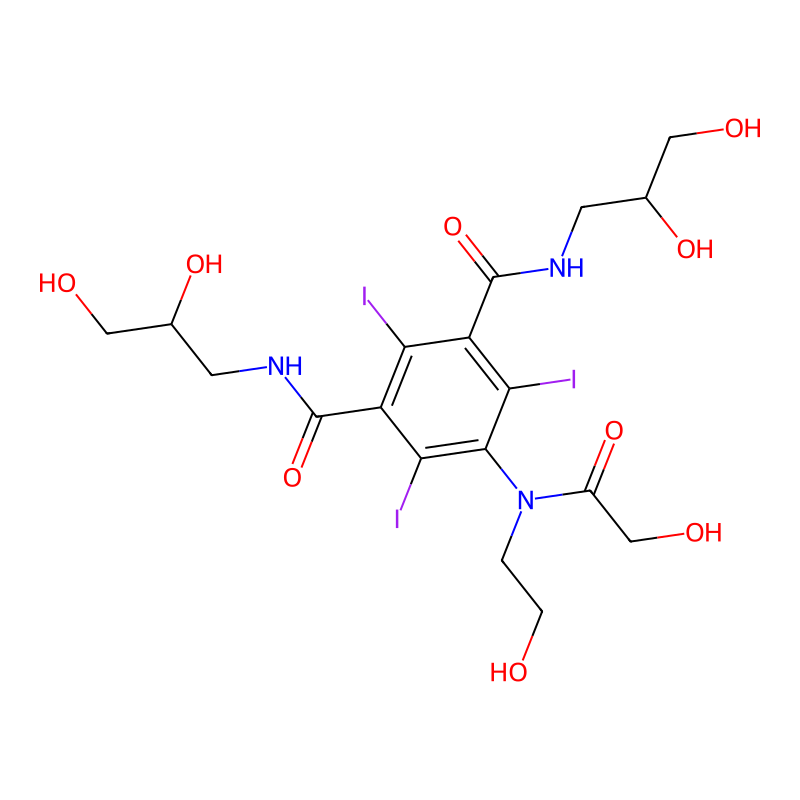}\label{fig:non_toxic}}
\hfill
\subfloat[Toxic (failed clinical trials)]{\includegraphics[width=0.48\columnwidth]{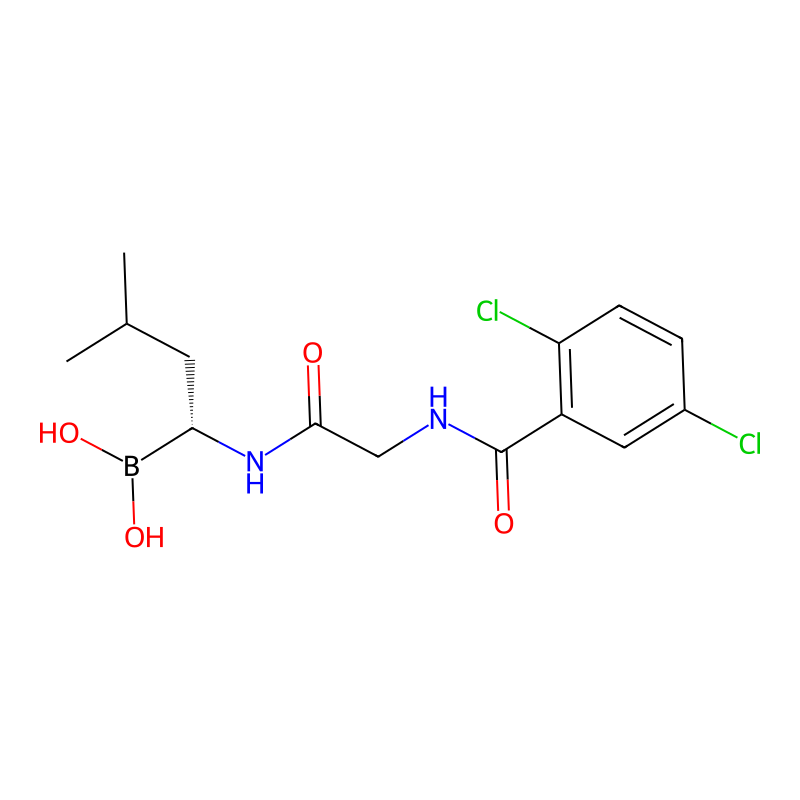}\label{fig:toxic}}
\caption{ClinTox examples: structurally similar molecules with opposite clinical outcomes.}
\label{fig:dataset_examples}
\end{figure}

ClinTox~\cite{ref_moleculenet} comprises 1,480 drug-like molecules with binary clinical toxicity labels (CT\_TOX task), exhibiting pronounced class imbalance (11.5:1). We apply scaffold-based splitting~\cite{ref_scaffold} (seed 42, 80/10/10) to evaluate generalization to novel molecular scaffolds. Fig.~\ref{fig:dataset_examples} illustrates that structurally similar molecules can have markedly different clinical outcomes.

We additionally evaluate on Tox21~\cite{ref_moleculenet}, comprising 7,831 molecules across 12 binary assays (nuclear receptors and stress response pathways) with up to 30\% missing annotations per task-handled via masked focal loss. The same scaffold split (seed 42, 80/10/10) yields 6,264/783/784 training/validation/test molecules; performance is mean AUC-ROC and mean PR-AUC across all 12 tasks.

\subsection{Data Processing}

\subsubsection{SMILES Sequence Processing}
For SMILESGNN, a custom tokenizer decomposes SMILES into chemical units (vocabulary 69, derived from the ClinTox training set~\cite{ref_smiles}). For SMILESGNN-PT, ChemBERTa-2 WordPiece subword tokenization is used. Both variants pad or truncate to 128 tokens.

\subsubsection{Molecular Graph Processing}
Atom nodes use 25 features (atomic number, formal charge, hybridization, stereochemistry, ring membership, aromaticity); edges use 17 features (bond type, direction, ring membership, conjugation, stereochemistry).

\subsection{SMILESGNN Architecture}

The overall architecture is illustrated in Fig.~\ref{fig:architecture}.

\subsubsection{SMILES Encoder}
The SMILES encoder employs a Transformer architecture with embedding dimension $d_{\text{model}} = 96$ and vocabulary size 69. Token embeddings are followed by learnable positional encodings. Two Transformer encoder layers, each with 4 attention heads and feedforward dimension 192, process the sequences with dropout 0.4. Masked mean pooling aggregates the sequence into $\mathbf{h}_{\text{SMILES}} \in \mathbb{R}^{96}$.

\subsubsection{Graph Encoder}
The graph encoder uses GATv2~\cite{ref_gatv2}. Atom features (25 dimensions) are projected to hidden dimension 96; edge features are embedded when available. Three GATv2 layers with 4 attention heads perform message passing with LayerNorm, ReLU, residual connections, and dropout 0.4. Jumping Knowledge connections~\cite{ref_jk} concatenate all layer outputs to produce 288-dim node representations. Mean-Max pooling generates $\mathbf{h}_{\text{graph}} \in \mathbb{R}^{576}$.

\subsubsection{Attention-Based Fusion Module}
The fusion module combines SMILES and graph representations through multi-head cross-attention~\cite{ref_attention} with 4 heads. The graph representation (576 dimensions) is projected to match the SMILES dimension (96); the SMILES representation serves as query, the projected graph as key and value. The attended output is concatenated with the original SMILES representation, yielding $\mathbf{h}_{\text{fused}} \in \mathbb{R}^{192}$, allowing the fused representation to condition sequence features on graph features.

\subsubsection{Predictor Head}
A four-layer MLP maps the fused 192-dimensional representation to a binary toxicity prediction: $192 \rightarrow 192 \rightarrow 96 \rightarrow 48 \rightarrow 1$. Each hidden layer includes BatchNorm, ReLU, and dropout (0.4).

\subsection{SMILESGNN-PT Architecture}
\label{sec:smilesgnn_pt}
SMILESGNN-PT replaces the custom Transformer SMILES encoder with a ChemBERTa-2~\cite{ref_chemberta2} pretrained backbone (RoBERTa pretrained on 77M PubChem SMILES, hidden size 384), while the GATv2 graph encoder is kept identical. The graph representation (576 dimensions) is projected to 384 via a linear layer with LayerNorm. Cross-attention fusion operates at 384 dimensions (ChemBERTa-2 mean-pool as query, projected graph as key/value), yielding $\mathbf{h}_{\text{fused}} \in \mathbb{R}^{768}$ by concatenation. The MLP predictor maps $768 \!\rightarrow\! 384 \!\rightarrow\! 192 \!\rightarrow\! 96 \!\rightarrow\! \text{output\_dim}$. SMILESGNN-PT has 4.7M total parameters (backbone 3.4M, head 1.3M), trained with a two-group AdamW ($\text{lr}_{\text{bb}}\!=\!10^{-4}$, $\text{lr}_{\text{hd}}\!=\!10^{-3}$), linear warmup, and gradient clipping (max norm 1.0).

\subsection{Training Methodology}

\subsubsection{Loss Function}
To address severe class imbalance, we employ Focal Loss~\cite{ref_focal}:
\begin{equation}
\mathcal{L}_{\text{focal}} = -\alpha (1 - p_t)^{\gamma} \log(p_t)
\end{equation}
where $p_t$ is the predicted probability for the true class, $\alpha = 0.25$, and $\gamma = 2.0$. For ClinTox (binary), this loss is applied directly. For Tox21 (multi-task), we use a masked variant that computes focal loss only over labelled (compound, assay) pairs, ignoring missing annotations.

\subsubsection{Optimization and Regularization}
SMILESGNN uses AdamW~\cite{ref_adamw} ($\text{lr}=5\!\times\!10^{-4}$, weight decay $10^{-4}$), batch size 32, up to 100 (ClinTox) or 60 (Tox21) epochs, early stopping on validation F1 / mean AUC-ROC (patience 20 / 10). SMILESGNN-PT uses the two-group AdamW from Section~\ref{sec:smilesgnn_pt}, batch size 16, otherwise identical. A weighted sampler, dropout (0.3-0.4), and batch normalization provide regularization.

\subsubsection{Evaluation Metrics}
Models are evaluated on the held-out test set using AUC-ROC, AUPRC, F1, and accuracy.

\section{Experiments and Results}
We evaluate SMILESGNN and SMILESGNN-PT against eight baselines on ClinTox (scaffold split, 1,184 train / 148 test). Single-modality baselines: Baseline MLP~\cite{ref_ecfp}, Pretrained GIN~\cite{ref_pretrained_gin}, GIN~\cite{ref_gin}, GATv2$^*$~\cite{ref_gatv2}, DMPNN~\cite{ref_dmpnn}, AttentiveFP~\cite{ref_attentivefp}, and SMILESTransformer$^*$~\cite{ref_molecular_transformer}. Multimodal concat baseline: CB2-GATv2$^\dagger$ (see Table~\ref{tab:results} caption), trained under identical conditions. SMILESGNN-PT$^\ddagger$ (Section~\ref{sec:smilesgnn_pt}) uses the same ChemBERTa-2 backbone as CB2-GATv2$^\dagger$ with cross-attention instead of concatenation, directly isolating the fusion strategy. Models marked $^*$ serve as single-modality ablation components.

\subsection{Overall Performance}

\begin{table*}[t]
\caption{Performance on ClinTox test set (scaffold split, 148 samples). \textbf{Bold}: proposed method.
$^*$Single-modality ablation components of SMILESGNN.
$^\dagger$Concat-fusion baseline: CB2-GATv2 pairs ChemBERTa-2~\cite{ref_chemberta2} (3.4M params) with GATv2, trained under identical conditions.
$^\ddagger$SMILESGNN-PT: same ChemBERTa-2 backbone as CB2-GATv2 with cross-attention fusion (4.7M params total).
All models trained under identical conditions (focal loss, balanced sampling). Values: mean$\pm$std, 3 seeds.}
\label{tab:results}
\setlength{\tabcolsep}{3pt}
\begin{center}
\small
\begin{tabular}{lcccc}
\toprule
\textbf{Model} & \textbf{AUC-ROC} & \textbf{Acc.} & \textbf{F1} & \textbf{AUPRC} \\
\midrule
\rowcolor{groupbg}
\multicolumn{5}{l}{\textit{Single-modality baselines}} \\
Pretrained GIN~\cite{ref_pretrained_gin}        & $0.722{\pm}0.024$ & $0.894{\pm}0.014$ & $0.170{\pm}0.101$ & $0.189{\pm}0.027$ \\
Baseline MLP~\cite{ref_ecfp}                    & $0.769{\pm}0.038$ & $0.914{\pm}0.017$ & $0.368{\pm}0.018$ & $0.390{\pm}0.087$ \\
GIN~\cite{ref_gin}                              & $0.815{\pm}0.056$ & $0.914{\pm}0.004$ & $0.423{\pm}0.038$ & $0.420{\pm}0.028$ \\
GATv2$^*$~\cite{ref_gatv2}                     & $0.819{\pm}0.025$ & $0.894{\pm}0.096$ & $0.424{\pm}0.162$ & $0.395{\pm}0.097$ \\
AttentiveFP~\cite{ref_attentivefp}              & $0.828{\pm}0.028$ & $0.849{\pm}0.026$ & $0.267{\pm}0.036$ & $0.368{\pm}0.025$ \\
DMPNN~\cite{ref_dmpnn}                          & $0.893{\pm}0.011$ & $0.919{\pm}0.012$ & $0.499{\pm}0.053$ & $0.453{\pm}0.018$ \\
SMILESTransformer$^*$~\cite{ref_molecular_transformer} & $0.995{\pm}0.003$ & $0.986{\pm}0.007$ & $0.896{\pm}0.053$ & $0.961{\pm}0.017$ \\
\midrule
\rowcolor{groupbg}
\multicolumn{5}{l}{\textit{Multimodal baselines — concat fusion ($^\dagger$)}} \\
CB2-GATv2$^\dagger$~\cite{ref_chemberta2}       & $0.987{\pm}0.018$ & $0.986{\pm}0.007$ & $0.890{\pm}0.063$ & $0.953{\pm}0.039$ \\
\midrule
\rowcolor{groupbg}
\multicolumn{5}{l}{\textit{Proposed methods}} \\
SMILESGNN-PT$^\ddagger$                         & $0.979{\pm}0.014$ & $0.982{\pm}0.004$ & $0.871{\pm}0.025$ & $0.930{\pm}0.013$ \\
\textbf{SMILESGNN}                              & $\mathbf{0.987{\pm}0.012}$ & $\mathbf{0.986{\pm}0.007}$ & $\mathbf{0.906{\pm}0.039}$ & $\mathbf{0.946{\pm}0.028}$ \\
\bottomrule
\end{tabular}
\end{center}
\end{table*}

Table~\ref{tab:results} shows that when all models are trained under identical conditions, the SMILES encoder alone (SMILESTransformer$^*$, $0.995\!\pm\!0.003$ AUC-ROC) is a strong single-modality baseline. All five graph-based baselines are substantially weaker (best: DMPNN $0.893\!\pm\!0.011$), confirming the challenge of scaffold-based graph generalization on ClinTox.

SMILESGNN achieves F1 $0.906\!\pm\!0.039$, exceeding SMILESTransformer$^*$ (F1 $0.896\!\pm\!0.053$) in minority-class identification while reducing variance, and matching CB2-GATv2$^\dagger$ (F1 $0.890\!\pm\!0.063$). In AUC-ROC ($0.987\!\pm\!0.012$), SMILESGNN is competitive with SMILESTransformer$^*$ within two standard deviations; the graph pathway adds GNNExplainer interpretability rather than a large accuracy gain on this small dataset (1,184 training molecules).

Replacing concat fusion with cross-attention under the identical ChemBERTa-2 backbone yields AUC-ROC $0.979\!\pm\!0.014$-within 0.008 of CB2-GATv2$^\dagger$ ($0.987\!\pm\!0.018$), with overlapping confidence intervals-confirming cross-attention is competitive. SMILESGNN outperforms SMILESGNN-PT on ClinTox (limited data favours the lightweight custom encoder); this advantage reverses on Tox21's 6,264 training samples ($+0.030$ AUC-ROC for SMILESGNN-PT).

\begin{figure*}[t]
\centering
\subfloat[ROC Curves]{\includegraphics[width=0.47\textwidth]{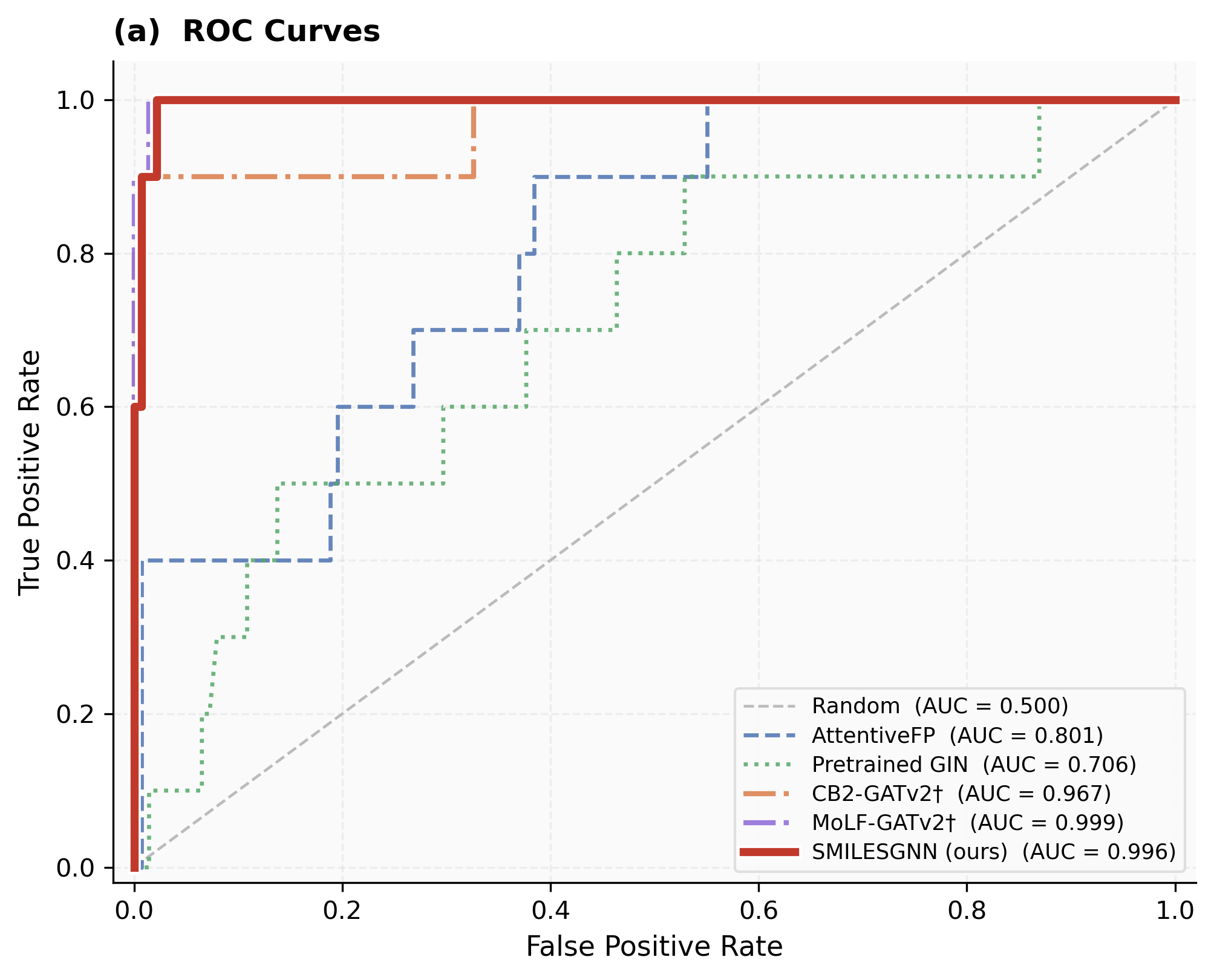}\label{fig:roc}}
\hfill
\subfloat[Precision-Recall Curves]{\includegraphics[width=0.47\textwidth]{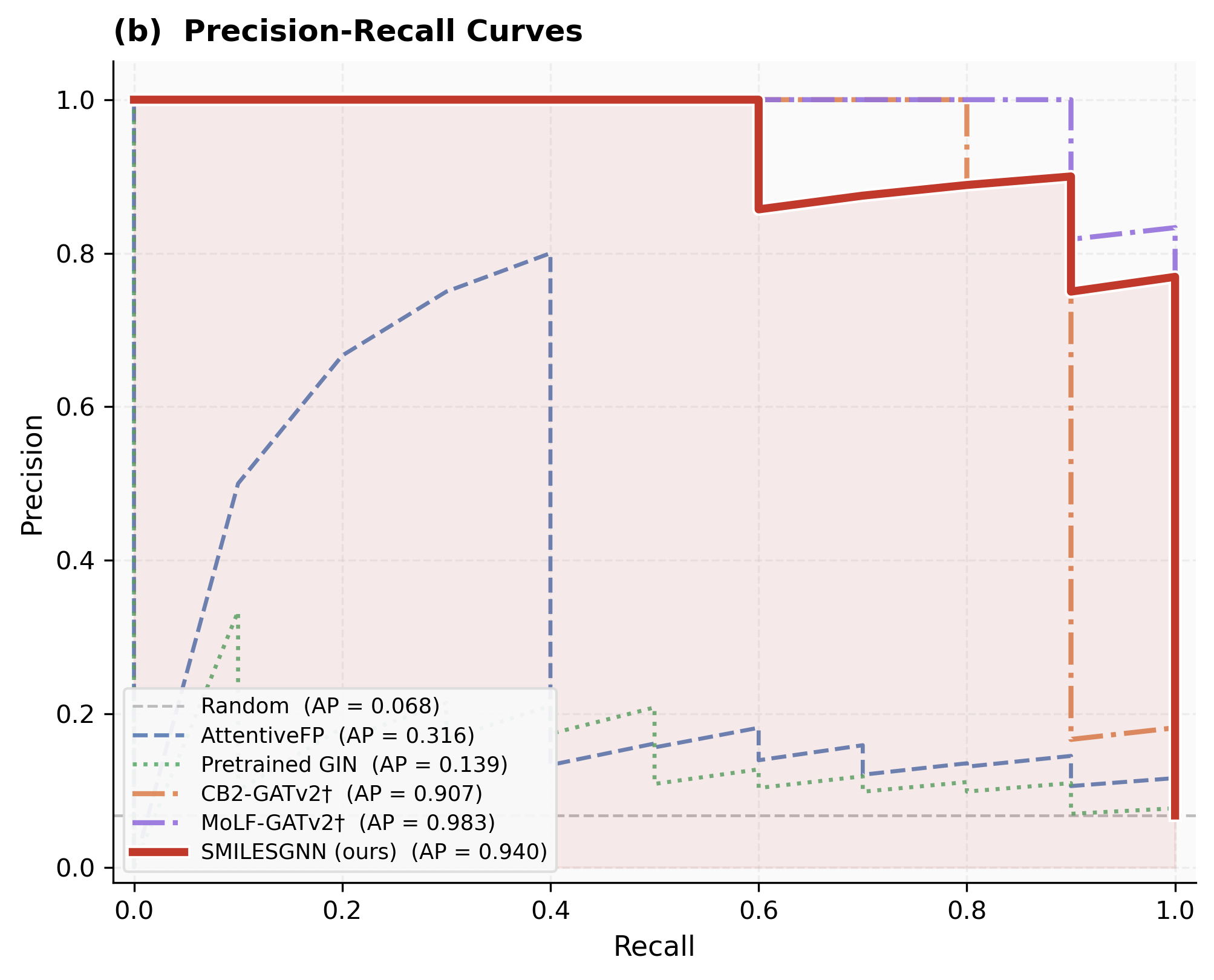}\label{fig:pr}}
\caption{Performance curves on the ClinTox test set (seed 42) for representative models: single-modality baselines (AttentiveFP, Pretrained GIN), multimodal concat baseline (CB2-GATv2$^\dagger$), and the proposed SMILESGNN. PR curves are more informative for imbalanced datasets; SMILESGNN (AUPRC 0.967 on this run) outperforms all graph-based baselines and matches the concat-fusion baseline.}
\label{fig:performance_curves}
\end{figure*}

Fig.~\ref{fig:performance_curves} shows ROC and Precision-Recall curves for representative models (seed 42). PR curves confirm SMILESGNN (mean AUPRC $0.946\!\pm\!0.028$) substantially outperforms graph-based single-modality baselines, validating multimodal fusion for imbalanced toxicity prediction~\cite{ref_pr_roc}.

\begin{figure}[htbp]
\centerline{\includegraphics[width=\columnwidth]{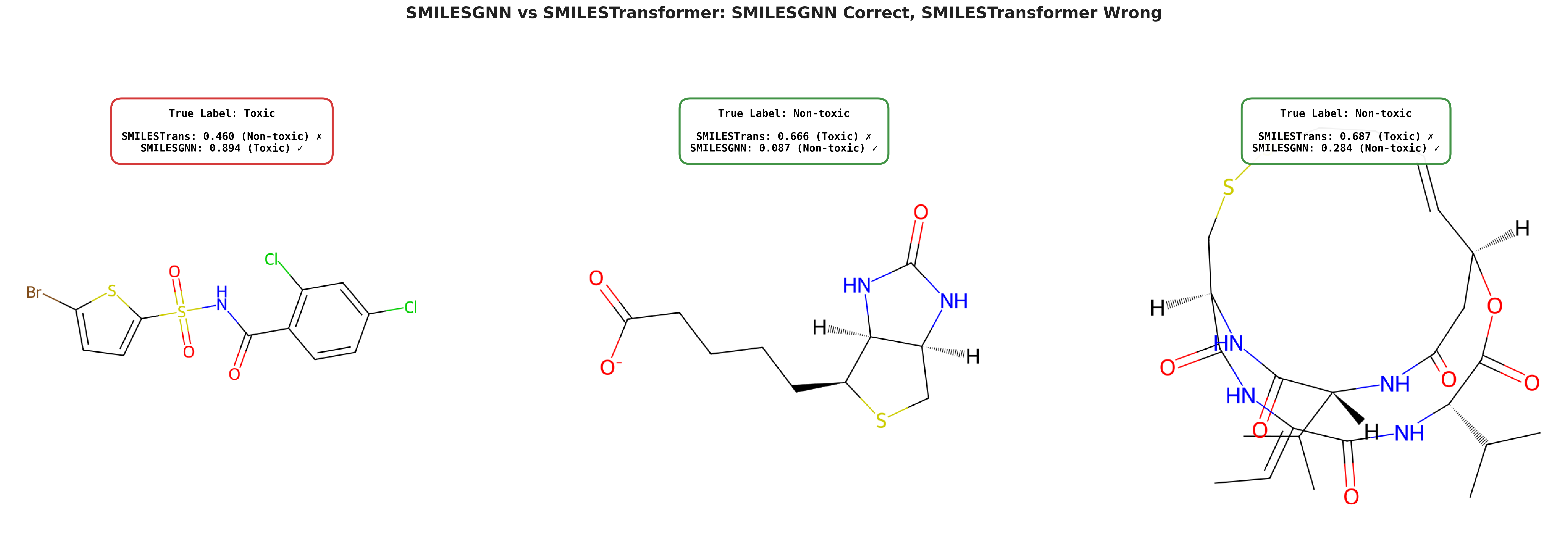}}
\caption{Three compounds correctly classified by SMILESGNN but misclassified by SMILESTransformer: one missed toxic detection (red border) and two corrected false positives (green border), illustrating graph encoder contributions beyond the SMILES sequence.}
\label{fig:smilesgnn_wins}
\end{figure}

We compare SMILESGNN with SMILESTransformer to isolate the graph contribution. Both models agree on 142 out of 148 test samples; for 3 disagreements favouring SMILESGNN (Fig.~\ref{fig:smilesgnn_wins}), it correctly identifies 1 missed toxic compound and avoids 2 false-positive errors, demonstrating that the graph encoder captures structural signals not apparent from the SMILES sequence alone.

\subsection{Fusion Mechanism Ablation}
To quantify the contribution of each encoder pathway, we compare three configurations using the same backbone components and training conditions.

\begin{table}[t]
\caption{Encoder contribution ablation on ClinTox. Equal-capacity projection ($576\!\to\!96$) applied before cross-attention fusion. $^*$Same encoders as SMILESGNN. \textbf{Bold}: proposed method. Values: mean$\pm$std, 3 seeds.}
\label{tab:ablation}
\begin{center}
\setlength{\tabcolsep}{1pt}
\footnotesize
\begin{tabular}{lccc}
\toprule
\textbf{Configuration} & \textbf{AUC-ROC} & \textbf{F1} & \textbf{AUPRC} \\
\midrule
\rowcolor{groupbg}
\multicolumn{4}{l}{\textit{Single branch (no fusion)}} \\
GATv2 only$^*$               & $0.819{\pm}0.025$ & $0.424{\pm}0.162$ & $0.395{\pm}0.097$ \\
SMILESTrans. only$^*$        & $0.995{\pm}0.003$ & $0.896{\pm}0.053$ & $0.961{\pm}0.017$ \\
\midrule
\rowcolor{groupbg}
\multicolumn{4}{l}{\textit{Dual branch (cross-attention fusion)}} \\
\textbf{Cross-attn} & $\mathbf{0.987{\pm}0.012}$ & $\mathbf{0.906{\pm}0.039}$ & $\mathbf{0.946{\pm}0.028}$ \\
\bottomrule
\end{tabular}
\end{center}
\end{table}

Table~\ref{tab:ablation} reveals two findings: (1) GATv2 alone is limited and high-variance (F1 $0.424\!\pm\!0.162$), while SMILESTransformer alone is strong (F1 $0.896\!\pm\!0.053$, AUC-ROC $0.995\!\pm\!0.003$); (2) cross-attention fusion achieves F1 $0.906\!\pm\!0.039$-slightly higher than SMILESTransformer alone with lower variance-suggesting that the graph pathway contributes complementary structural signals for minority-class identification. Cross-attention fusion additionally retains an explicit graph branch for GNNExplainer analysis (Section~\ref{sec:exp:interp}).

\subsection{Tox21 Multi-Task Evaluation}

\begin{table}[t]
\caption{Mean AUC-ROC and PR-AUC on Tox21 (12 tasks, scaffold split). $^\dagger$Same conditions as Table~\ref{tab:results}. $^\ddagger$ChemBERTa-2 + cross-attn + GATv2 (4.7M). \textbf{Bold}: best proposed. Values are mean$\pm$std over 3 seeds.}
\label{tab:tox21}
\begin{center}
\resizebox{\columnwidth}{!}{%
\begin{tabular}{l>{\raggedright\arraybackslash}p{2.0cm}cc}
\toprule
\textbf{Model} & \textbf{Modality} & \textbf{AUC-ROC} & \textbf{PR-AUC} \\
\midrule
\rowcolor{groupbg}
\multicolumn{4}{l}{\textit{Single-modality baselines}} \\
ECFP4+XGBoost~\cite{ref_ecfp}        & Fingerprint        & $0.706{\pm}0.001$ & $0.288{\pm}0.007$ \\
AttentiveFP~\cite{ref_attentivefp}    & Graph              & $0.740{\pm}0.008$ & $0.318{\pm}0.002$ \\
ChemBERTa-2~\cite{ref_chemberta2}     & Seq. (pretrained)  & $0.752{\pm}0.009$ & $0.333{\pm}0.024$ \\
\midrule
\rowcolor{groupbg}
\multicolumn{4}{l}{\textit{Multimodal baselines — concat fusion ($^\dagger$)}} \\
CB2-GATv2$^\dagger$~\cite{ref_chemberta2}  & SMILES + Graph  & $0.751{\pm}0.012$ & $0.351{\pm}0.015$ \\
\midrule
\rowcolor{groupbg}
\multicolumn{4}{l}{\textit{Proposed methods}} \\
SMILESGNN                             & SMILES + Graph      & $0.720{\pm}0.008$ & $0.279{\pm}0.010$ \\
\textbf{SMILESGNN-PT$^\ddagger$}      & \textbf{SMILES + Graph} & $\mathbf{0.750{\pm}0.002}$ & $\mathbf{0.334{\pm}0.012}$ \\
\bottomrule
\end{tabular}%
}
\end{center}
\end{table}

On Tox21, SMILESGNN-PT$^\ddagger$ ($0.750\!\pm\!0.002$) outperforms SMILESGNN ($0.720\!\pm\!0.008$) by $+0.030$, confirming the benefit of pretrained representations. SMILESGNN-PT$^\ddagger$ matches ChemBERTa-2 alone ($0.752\!\pm\!0.009$) and CB2-GATv2$^\dagger$ ($0.751\!\pm\!0.012$)-overlapping confidence intervals-directly demonstrating cross-attention is competitive with concatenation under the same backbone.

\subsection{Graph Representation Analysis}
\label{sec:exp:interp}

\begin{figure*}[t]
\centerline{\includegraphics[width=0.7\textwidth]{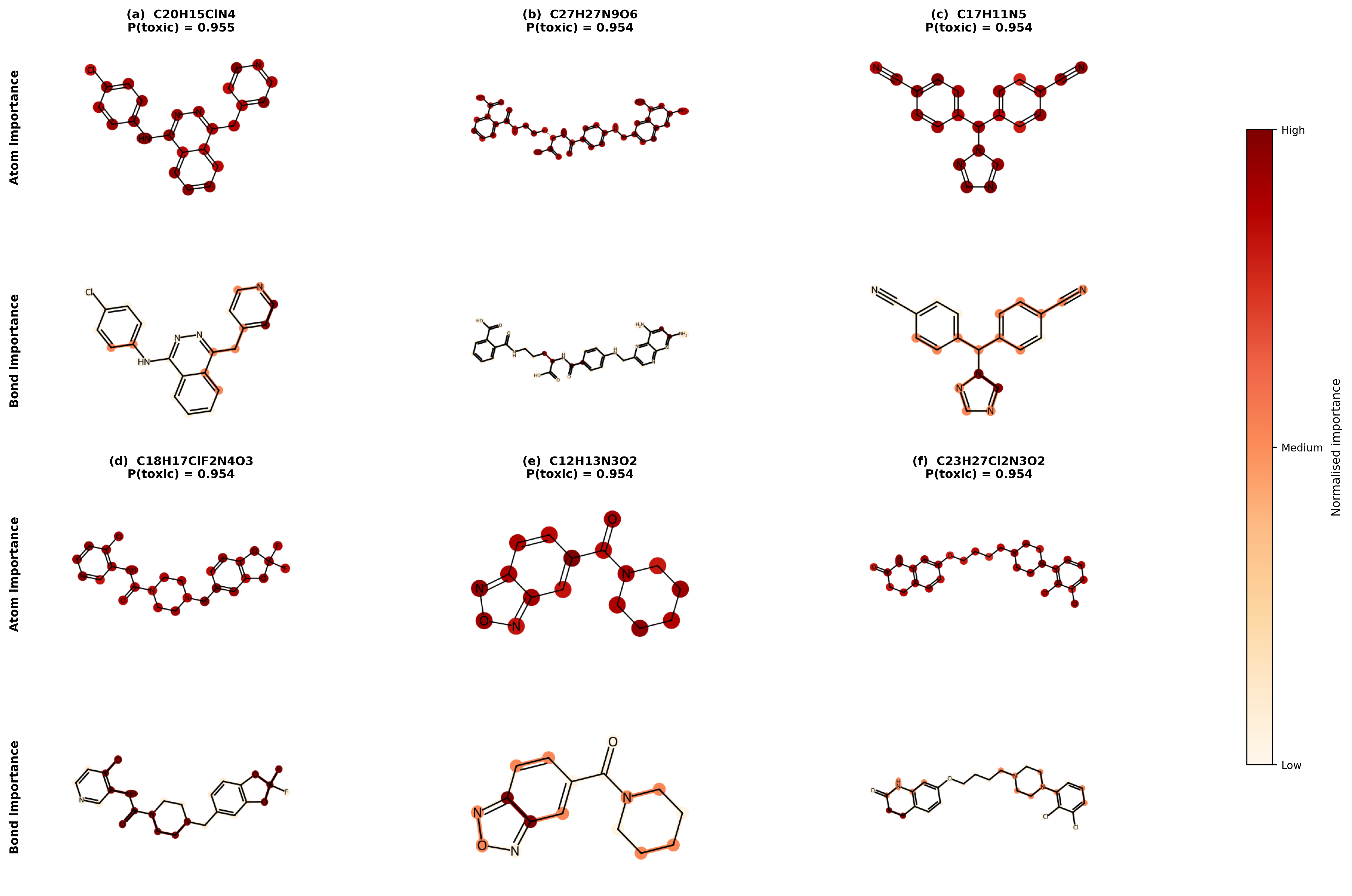}}
\caption{GNNExplainer~\cite{ref_gnnexplainer} interpretability analysis of the six highest-confidence ClinTox toxic predictions ($P_{\text{tox}}\!\geq\!0.954$). Two blocks of three molecules each; \textit{Row 1 (Atom importance)} and \textit{Row 2 (Bond importance)} per block; darker red indicates higher learned mask importance for the toxic classification. SMILES embeddings are frozen to isolate graph-attributed explanations.}
\label{fig:gnnexplainer}
\end{figure*}

To demonstrate model interpretability, we apply GNNExplainer~\cite{ref_gnnexplainer} to the six highest-confidence toxic predictions from the ClinTox test set (Fig.~\ref{fig:gnnexplainer}). GNNExplainer optimises per-atom and per-bond soft masks over the GATv2 pathway to identify compact subgraphs associated with the toxic classification. Because the graph encoder remains part of the fused predictor, these masks provide a graph-branch view of the learned decision process; we interpret them as qualitative, model-specific evidence rather than a formal proof of causality.

Three chemically interpretable patterns recur across all six molecules. For halogenated aromatics (a, d), atom and bond importance concentrates on halogenated phenyl rings and conjugated bridges, consistent with cytochrome P450-mediated activation — a well-documented cytotoxicity mechanism. For reactive heterocycles (c, e), importance peaks at triazole and isoxazole rings, corresponding to structural alerts for metabolic nitrile hydrolysis and ring-opening electrophilic reactivity. For extended scaffolds (b, f), importance is more diffuse but consistently localises at central amide linkers and terminal halogenated rings.

\section{Conclusion}
This paper introduced SMILESGNN, a multimodal toxicity prediction framework that combines SMILES sequence modeling, graph message passing, and cross-attention fusion under class-imbalance-aware training. Its central contribution is to retain an explicit molecular graph pathway while preserving competitive predictive performance, allowing benchmark evaluation to be complemented by GNNExplainer-based inspection of toxicophore-like substructures. The pretrained variant, SMILESGNN-PT, further suggests that the same fusion strategy can be transferred to a ChemBERTa-2 backbone and remain competitive with a comparable concat-fusion design.
However, ClinTox is small and highly imbalanced, the explanations are qualitative rather than causal, and the current evaluation lacks stronger stress tests such as graph perturbation, calibration analysis, and external clinical validation.
Future work will therefore focus on validating the graph branch more rigorously through perturbation-based ablations, graph-level pretraining, uncertainty calibration, and evaluation on larger toxicity benchmarks and external clinical datasets. We also plan to study task-adaptive loss weighting and explanation stability so that cross-attention fusion can better support both robust prediction and chemically meaningful interpretation.

\bibliographystyle{IEEEtran}
\bibliography{main}

\end{document}